\documentclass{isprs} 
\usepackage{subfigure}
\usepackage{setspace}
\usepackage{geometry} 
\usepackage{epstopdf}
\usepackage[labelsep=period]{caption}  
\usepackage[british]{babel} 
\usepackage[hang]{footmisc}
\usepackage{amsmath}
\usepackage{booktabs} 
\usepackage{graphicx} 
\usepackage[none]{hyphenat} 
\usepackage{tabularx}
\usepackage{chngcntr} 
\usepackage{hyperref}
\usepackage{pifont}

\begin{document}

\title{Accurate Point Measurement in 3DGS - A New Alternative to Traditional Stereoscopic-View Based Measurements}
\date{}


\author{
Deyan Deng$^{1}$, Rongjun Qin$^{1,2}\thanks{Corresponding author}$
}
\address{
\textsuperscript{1} Dept. of Civil, Environmental and Geodetic Engineering, The Ohio State University, Columbus OH, US \\
\textsuperscript{2} Dept. of Electrical and Computer Engineering, The Ohio State University, Columbus OH, US \\
\{deng.1069, qin.324\}@osu.edu
}



\abstract{

3D Gaussian Splatting (3DGS) has revolutionized real-time rendering with its state-of-the-art novel view synthesis, but it is often criticized by its useability in accurate geometric measurements. As compared multi-view stereo (MVS) based point clouds or mesh models, 3DGS rendered views present better visual quality and completeness, while the current methods for point measurements, still relies on stereoscopic measurement on stereo workstations, or direct measurements on the often-incomplete and inaccurate 3D point clouds or meshes. As a novel view synthesizer, 3DGS renders the exact source views, and interpolate in-between views to allow very smooth view transition. This feature allows  users to intuitively pick congruent points in different views while operating the 3DGS models. By triangulating these congruent points, One can precisely generate point measurements. This idea is similar to traditional stereoscopic measurement, while it is much less demanding, for example, it does not require a stereo workstation nor operators with biological sterescopic capability. Moreover, it allows users to measure on more than two views for higher measurement accuracy. We have implemented a web-based application to demonstrate this proof of concept (PoC). Using several aerial dataset captured by unmanned aerial vehicles (UAV), we show that our work implementing this PoC allows users to successfully perform highly accurate point measurements, achieving accuracy similar or beyond traditional stereoscopic measurements but using non-stereo workstations. Specifically, we show that our measured points significantly outperforms point measurements directly on the generated meshes. Quantitatively, our method achieves RMSEs in the 1-2 cm range on well-defined points, consistently outperforming direct mesh measurements. More critically, on challenging thin structures where mesh-based RMSE was 0.062 m, our method achieved 0.037 m. On sharp corners poorly reconstructed in the mesh, our method successfully measured all points with a 0.013 m RMSE, whereas the mesh-based method failed entirely. The source code and documentation for the proposed 3DGS measurement tool are open-source and available at: \url{https://github.com/GDAOSU/3dgs_measurement_tool}.
}


\maketitle
 
\sloppy

\section{Introduction}

3D point coordinate measurement is a core process in the surveying, mapping, and construction industries. Traditionally, this was dominated by \textit{in-situ} methods such as total stations and Global Navigation Satellite Systems (GNSS), which provide high-accuracy coordinates through direct field observation of discrete points. The evolution toward non-contact, remote sensing techniques—particularly photogrammetry and LiDAR—marked a significant paradigm shift \cite{Luhmann2013,Wehr1999,Vosselman2010,tongCrackParameterDetermination2025}.

The current practice in generating 3D models is largely dominated by automated dense image matching and multi-view stereo (MVS) algorithms \cite{Hirschmuller2008,Furukawa2010}. While this paradigm has significantly advanced applications such as large-scale surface modeling and true-orthophoto rectification, the resulting \textit{explicit geometric representations} (e.g., dense point clouds and 3D meshes) still suffer from measurement inaccuracies. The reliability of querying coordinates from these models is fundamentally constrained by the geometric fidelity and surface smoothing inherent in the reconstruction pipeline. Consequently, for applications demanding high-precision point measurements, the industry still often relies on traditional manual or stereoscopic approaches, which require specialized hardware and operator expertise \cite{Luhmann2013,McGlone2013}.

Recently, 3D Gaussian Splatting (3DGS) has emerged as a disruptive technology in both research and industry \cite{Kerbl2023,Li2025,Huang2025a,zhongAHighQualityUnderwater3D2025}. Building on breakthroughs in neural rendering, including NeRF-based scene representations \cite{Mildenhall2020} and neural point-based rendering \cite{Aliev2020}, 
3DGS provides high-fidelity, photorealistic renderings in real-time. However, unlike a mesh or point cloud, 3DGS does not encode an explicit surface; it represents a scene as a set of optimized 3D Gaussians designed for view synthesis. This presents a challenge for metrology, as a simple “pick” on a rendered image does not correspond to a well-defined geometric surface.

\begin{figure}
    \centering
    \includegraphics[width=1.0\linewidth]{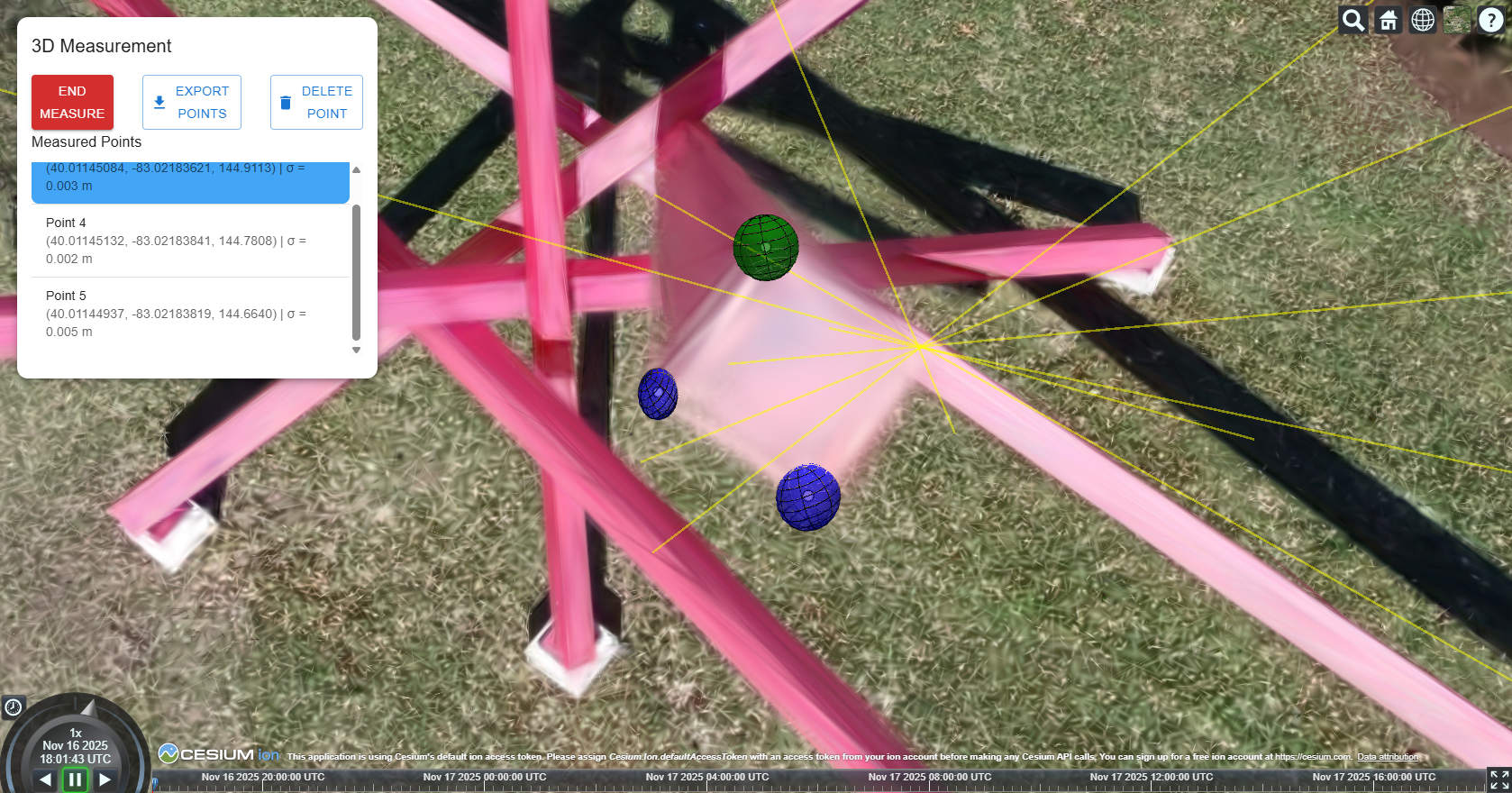}
    \caption{User interface of the web-based measurement application. The panel (left) lists the computed 3D points and their corresponding RMS error ($\sigma$). In the 3D view, multiple user-aimed rays (yellow lines) are used to compute the 3D point via spatial intersection. The final point is visualized as a 3D error ellipsoid (blue/green), providing immediate feedback on the measurement's precision.}
    \label{fig:Measurement UI}
\end{figure}

To bridge this gap, this study proposes a new alternative by developing and validating a web-based application that enables precise 3D point measurement from 3DGS renderings. Our system is built on the Cesium platform \cite{CesiumJS}, streaming 3DGS results converted to 3D Tiles \cite{OGC2018}. We reintroduce the classical principle of stereo measurement—spatial intersection—but implement it in a flexible, modern visualization environment. The method computes a 3D coordinate as the least-squares intersection of multiple rays, following standard multi-view geometry formulations \cite{Hartley2003}. 
Figure~\ref{fig:Measurement UI} shows the user interface of the developed web application.

As an alternative to traditional stereoscopic-view based systems, our web-based implementation of spatial intersection offers several significant advantages over the classic methods it modernizes. First, it democratizes the measurement process by removing the reliance on specialized hardware, such as stereo glasses or 3D mice. Consequently, it no longer requires the operator to possess biological stereo viewing capabilities, making it accessible to a wider range of users. Second, it simplifies the workflow by directly leveraging the camera pose from the rendering engine (e.g., Cesium), eliminating the need for complex camera model calibration and pose format conversions between different software packages. Finally, our method is not limited to the two views of a stereo pair; it inherently supports a multi-ray ($N > 2$) intersection, which allows for a more robust least-squares adjustment (as discussed in Sec. 2.1) and can deliver higher precision than traditional two-ray stereo triangulation. 

To validate the accuracy and advantages of this approach, we compare our results against (1) direct measurements on a photogrammetric mesh model and (2) high-precision ground truth coordinates derived from tie-point triangulation. This analysis verifies that our web-based tool is an accurate and reliable method, combining the rendering fidelity of 3DGS with the metrological rigor of photogrammetry.

This work presents a measurement-oriented framework for 3DGS that enables accurate 3D point measurement within a modern, web-based visualization environment. By integrating classical forward intersection principles with 3DGS rendering, the proposed approach extends the use of 3DGS from visualization to quantitative measurement. The main contributions of this paper are threefold:
\begin{itemize}
    \item A web-based multi-ray spatial intersection method that enables accurate 3D point measurement directly from 3DGS renderings;
    \item An intuitive uncertainty-aware user interface that provides immediate feedback on measurement precision;
    \item A quantitative evaluation demonstrating accuracy comparable to established photogrammetric approaches and superior reliability relative to mesh-based measurements.
\end{itemize}

\section{Methodology}
Our methodology leverages the unique strength of 3DGS as a high-fidelity view exploration tool. Unlike static mesh models, 3DGS provides seamless, photorealistic novel view synthesis, allowing an operator to intuitively navigate a scene and inspect a target of interest from multiple perspectives. This smooth view transition, which replicates moving between the original dense source images, makes it straightforward for a user to identify the same semantic point, referred to as a congruent point, across several different views.

The core of our method is to then apply the classic, robust photogrammetric principle of spatial intersection to these user-identified congruent points. By collecting the viewing rays from multiple perspectives, we can compute a highly accurate 3D coordinate for the target, bypassing the geometric inaccuracies of an underlying mesh.

\subsection{Core Principle: 3D Point from Spatial Intersection}

In traditional photogrammetry, spatial intersection (or forward triangulation) computes the 3D coordinate of a point by finding the intersection of at least two 3D rays. Each ray is defined by an image measurement and its corresponding camera's pose. Our method applies this exact principle, using the 3DGS rendering as a high-fidelity guide for aiming these rays.

\subsubsection{Geometric and Algebraic Formulation}
A single 3D ray, $\mathbf{R}_i$, is defined by its origin $\mathbf{C}_i = (X_i, Y_i, Z_i)^T$ (the camera center) and a 3D unit direction vector $\mathbf{d}_i = (u_i, v_i, w_i)^T$. We seek to find the unknown 3D point $\mathbf{P} = (x, y, z)^T$ that lies at the intersection of $N$ such rays (where $N \ge 2$).

Geometrically, if the point $\mathbf{P}$ lies perfectly on the ray $\mathbf{R}_i$, then the vector from the camera center to the point, $(\mathbf{P} - \mathbf{C}_i)$, must be parallel to the ray's direction vector $\mathbf{d}_i$. In a perfect, noise-free system, the cross product of two parallel vectors is the zero vector:

\begin{equation}
    (\mathbf{P} - \mathbf{C}_i) \times \mathbf{d}_i = \mathbf{0}
\end{equation}

Let $\Delta\mathbf{P}_i = \mathbf{P} - \mathbf{C}_i = (x-X_i, y-Y_i, z-Z_i)^T$. Expanding the cross product yields three equations:

\begin{equation}
\begin{cases}
(y - Y_i)w_i - (z - Z_i)v_i = 0 \\
(z - Z_i)u_i - (x - X_i)w_i = 0 \\
(x - X_i)v_i - (y - Y_i)u_i = 0
\end{cases}
\end{equation}

Although there are three equations, only two are linearly independent. We can select any two to form a linear system for that ray. Choosing the first two and rearranging them to isolate the unknown parameters $(x, y, z)$ on one side, we get:

\begin{equation}
    \begin{cases}
    (0)x + (w_i)y + (-v_i)z = w_i Y_i - v_i Z_i \\
    (-w_i)x + (0)y + (u_i)z = -w_i X_i + u_i Z_i
    \end{cases}
\end{equation}

\subsubsection{Least-Squares Solution}
This system is for a single ray. In a real-world scenario, due to small aiming errors, the $N$ rays will not intersect at a single perfect point. We therefore formulate a linear least-squares problem, $\mathbf{Ax} = \mathbf{b}$, to find the point $\mathbf{P}$ that minimizes the sum of squared distances to all $N$ rays.

For each ray $i$, we define a $2 \times 3$ matrix $\mathbf{A}_i$ and a $2 \times 1$ vector $\mathbf{b}_i$:

\begin{equation}
\mathbf{A}_i =
\begin{bmatrix}
0 & w_i & -v_i \\
-w_i & 0 & u_i
\end{bmatrix}
, \quad
\mathbf{b}_i =
\begin{bmatrix}
w_i Y_i - v_i Z_i \\
-w_i X_i + u_i Z_i
\end{bmatrix}
\end{equation}

The full system for $N$ rays is constructed by stacking these individual matrices:

\begin{equation}
\mathbf{A} =
\begin{bmatrix}
\mathbf{A}_1 \\ \mathbf{A}_2 \\ \vdots \\ \mathbf{A}_N
\end{bmatrix}
, \quad
\mathbf{b} =
\begin{bmatrix}
\mathbf{b}_1 \\ \mathbf{b}_2 \\ \vdots \\ \mathbf{b}_N
\end{bmatrix}
, \quad
\mathbf{x} =
\begin{bmatrix}
x \\ y \\ z
\end{bmatrix}
\end{equation}

Here, $\mathbf{A}$ is a $2N \times 3$ matrix, $\mathbf{b}$ is a $2N \times 1$ vector, and $\mathbf{x}$ is the $3 \times 1$ vector of the unknown 3D coordinates. This overdetermined system is robustly solved for $\mathbf{x}$ using the normal equation:

\begin{equation}
\mathbf{x} = (\mathbf{A}^T\mathbf{A})^{-1}\mathbf{A}^T\mathbf{b}
\label{eq:lsq}
\end{equation}

\subsubsection{Uncertainty Assessment}
After solving for $\mathbf{x}$, the quality of the intersection can be assessed. The vector of residuals $\mathbf{v}$ is computed as $\mathbf{v} = \mathbf{Ax} - \mathbf{b}$. The \textit{a posteriori} variance of unit weight, $\hat{\sigma}_0^2$, is then given by:

\begin{equation}
\hat{\sigma}_0^2 = \frac{\mathbf{v}^T\mathbf{v}}{r}
\label{eq:variance}
\end{equation}

where $r$ is the redundancy of the system (degrees of freedom), calculated as $r = 2N - 3$ (for $N$ rays, each contributing 2 equations, and 3 unknown parameters).

The Root Mean Square (RMS) of the residuals, which represents the average geometric distance of the rays from the computed point, is $\text{RMS} = \sqrt{\hat{\sigma}_0^2}$. This RMS value serves as a crucial uncertainty indicator for each measurement. Furthermore, the covariance matrix for the estimated point $\mathbf{P}$ is:

\begin{equation}
\mathbf{Q}_{\hat{x}\hat{x}} = \hat{\sigma}_0^2 (\mathbf{A}^T\mathbf{A})^{-1}
\end{equation}

The diagonal elements of $\mathbf{Q}_{\hat{x}\hat{x}}$ provide the variances ($\sigma_x^2, \sigma_y^2, \sigma_z^2$) of the computed coordinates, allowing for a rigorous, quantitative assessment of the measurement's precision.

Although an arbitrary number of synthetic observations can be generated, they do not increase the effective degrees of freedom beyond those provided by the original images. Increasing the number of sampled points primarily improves numerical stability rather than introducing new independent information.

\subsection{Algorithm Flow for 3DGS-based Measurement}
The 3DGS model provides the two components necessary for this method: (1) a high-fidelity rendered image and (2) the precise camera pose for that rendering. Our method overcomes the lack of an explicit surface by using the 3DGS rendering purely as a high-fidelity \textit{visual reference} for aiming the measurement rays. The algorithm flow for a single point measurement is as follows:

\begin{quote}
\textbf{Algorithm 1: Multi-Ray Spatial Intersection from 3DGS}
\begin{enumerate}
    \item \textbf{Initialize:} Create an empty list $L_{rays}$ to store measurement rays.
    \item \textbf{Loop} for $i = 1$ to $N$ (where $N \ge 2$ is the number of views):
    \begin{itemize}
        \item Operator navigates the 3D viewer to a desired viewpoint $V_i$.
        \item Operator selects a pixel $p_i$ corresponding to the congruent point.
        \item Get current camera pose $C_i$ (origin) and direction $D_i$ (from $C_i$ through $p_i$) from the rendering engine.
        \item Define ray $\mathbf{R}_i = (\mathbf{C}_i, \mathbf{d}_i)$, where $\mathbf{C}_i$ is the camera origin and $\mathbf{d}_i$ is the unit direction vector.
        \item Add $\mathbf{R}_i$ to $L_{rays}$.
    \end{itemize}
    \item \textbf{Compute:} Construct the $\mathbf{A}$ matrix and $\mathbf{b}$ vector by stacking the components ($\mathbf{A}_i, \mathbf{b}_i$) from each ray $\mathbf{R}_i \in L_{rays}$ (as defined in Sec. 2.1.2).
    \item \textbf{Solve:} Calculate the 3D point $\mathbf{x} = (x, y, z)^T$ using the least-squares solution: $\mathbf{x} = (\mathbf{A}^T\mathbf{A})^{-1}\mathbf{A}^T\mathbf{b}$.
    \item \textbf{Assess:} (Optional) Compute residuals $\mathbf{v} = \mathbf{Ax} - \mathbf{b}$ and uncertainty metrics as defined in Sec. 2.1.3.
    \item \textbf{Return:} The 3D coordinate $\mathbf{x}$, standard deviation $\hat{\sigma_0}$ and covariance matrix $\mathbf{Q}_{\hat{x}\hat{x}}$.
\end{enumerate}
\end{quote}

The resulting 3D coordinate $\mathbf{x}$, along with its computed uncertainty (e.g., the full covariance matrix $\mathbf{Q}_{\hat{x}\hat{x}}$ and the derived standard deviations), is then recorded in a list of all measured points for subsequent use.

\subsection{Web-Based PoC Implementation}

To validate this methodology, we implemented the procedure described in Algorithm 1 as a web-based proof-of-concept (PoC) application. This implementation allows operators to perform measurements from any standard web browser. The client-side application is built directly upon the \textbf{Cesium.js platform \cite{CesiumJS}}, which provides the core 3D globe, all camera control APIs, and the high-fidelity 3DGS rendering engine. The system's technical architecture, which handles secure data streaming and computation, is briefly summarized in Table \ref{tab:implementation}.

\begin{table}[h]
\centering
\caption{System Implementation Architecture}
\label{tab:implementation}
\small
\begin{tabularx}{\columnwidth}{lX} 
\toprule
\textbf{Component} & \textbf{Technology \& Role} \\
\midrule
\textbf{Data Storage} & Amazon S3. Hosts 3DGS models converted to OGC 3D Tiles. \\
\textbf{Backend} & Node.js / Express.js. Serves the client application and acts as a secure gateway to stream 3D Tiles data from S3. \\
\textbf{Web Client} & \textbf{Cesium.js} / React. \textbf{Cesium.js} provides the core 3DGS rendering, 3D globe, and camera APIs. React manages the UI overlays. \\
\textbf{Computation} & JavaScript. Implements the client-side least-squares solution (Eq. \ref{eq:lsq}) within the client application. \\
\bottomrule
\end{tabularx} 
\end{table}

\section{Experiments and Analysis}

Our primary hypothesis is that 3DGS-based models, when queried with our proposed multi-ray intersection method, yield more accurate and robust 3D measurements than direct measurements on conventional mesh models. We further hypothesize that this advantage is most pronounced in challenging geometric areas—such as thin structures and sharp edges—where mesh representations are notoriously prone to smoothing or failure.

To test this, we designed a quantitative experiment to assess absolute accuracy against a high-precision ground truth. We selected three distinct, real-world UAV datasets (Fig. \ref{fig:datasets}) that embody these common challenges (e.g., reflective surfaces, thin poles, complex trusses). Success on these varied scenes provides strong evidence for the general applicability of our method.

\subsection{Experimental Setup}

\subsubsection{Datasets}

We tested this idea on three datasets captured with UAV imagery and processed using Bentley iTwin Capture Modeler. This software was used to generate both the 3DGS scenes and the conventional 3D textured meshes. 
\begin{itemize}
    \item \textbf{Dataset 1 (Steel Sculpture \footnote{The artwork is titled 'Out of Bounds'.}):} A white steel statue with complex, smooth, and reflective surfaces. It includes several thin, pole-like structures (e.g., traffic light poles, instruction poles). This dataset was used for accuracy and the advantage analysis.
    \item \textbf{Dataset 2 (Wood Sculpture \footnote{The artwork is titled 'PIVOT'.}):} A structure made of red wood and steel. It features sharp, rectangular corners at its base. This dataset was also used for accuracy and the advantage analysis.
    \item \textbf{Dataset 3 (Campus Building):} A building on campus featuring complex steel structures, fine trusses, and thin railings. This dataset was used for the accuracy assessment.
\end{itemize}

\begin{figure}
    \centering
    \includegraphics[width=1.0\linewidth]{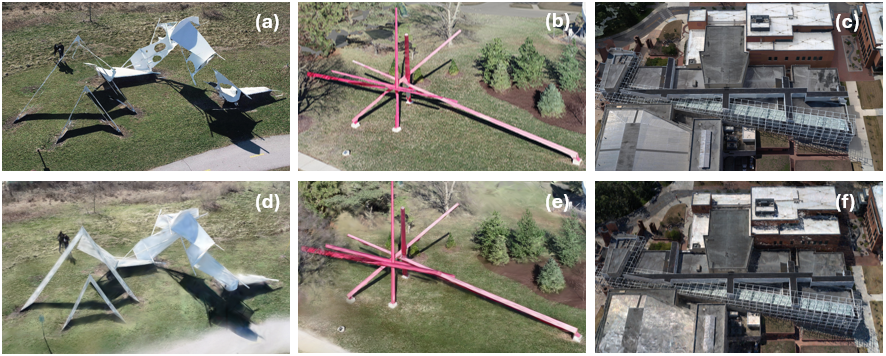}
    \caption{Datasets used for experiments. (a - c): mesh models for Dataset 1 - 3; (d - f): 3DGS models for Datasets 1 - 3.}
    \label{fig:datasets}
\end{figure}

Figure \ref{fig:datasets} shows mesh and 3DGS models for the three datasets used in this study.

\subsubsection{Ground Truth Generation}

To establish a high-precision ground truth, we defined a set of Validation Points (VPs) for each scene. These VPs are high-accuracy 3D coordinates derived from multi-ray photogrammetric triangulation, and they serve as our "gold standard" for error analysis. This process was conducted \textit{after} a successful aerial triangulation (AT) of the entire image block using the tools within iTwin Capture Modeler.

For each of the 20 desired VPs, we used the "Survey" tools to "Add Tie Point." This initiated a multi-view measurement workflow. The operator manually identified the target feature (e.g., a building corner) on one image and then selected it on multiple corresponding images. To ensure the highest accuracy, the operator zoomed in significantly on each image to achieve precise, sub-pixel pointing.

\begin{figure}[h]
    \centering
    \includegraphics[width=1.0\linewidth]{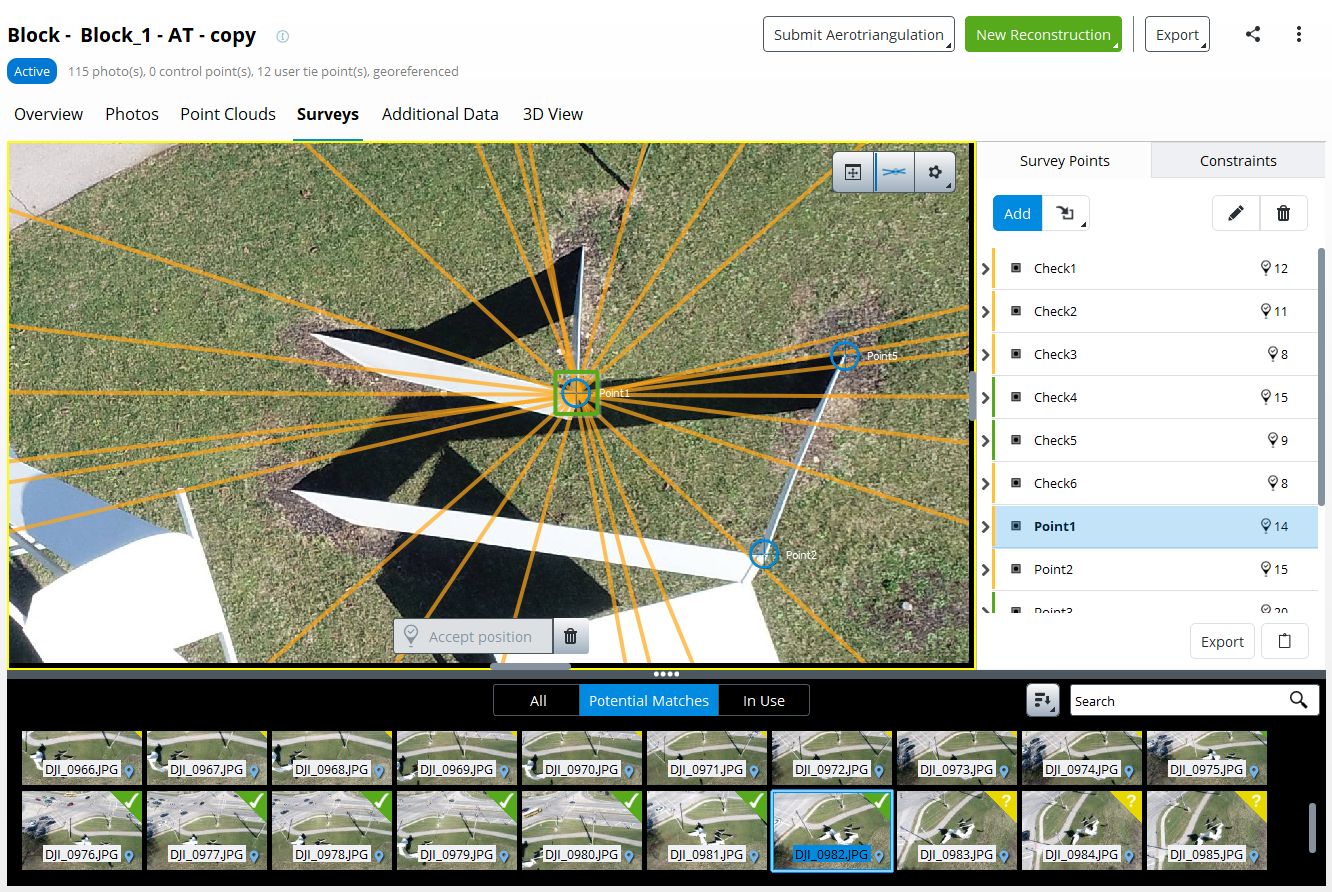}
    \caption{Ground truth generation in iTwin Capture Modeler \protect\cite{Bentley2024}. A single validation point (VP) is precisely measured across multiple source images. The system then computes its 3D coordinate via rigorous spatial intersection, which serves as the high-accuracy ground truth for our validation.}
    \label{fig:ground_truth}
\end{figure}

Each VP was measured in a minimum of five images (and often more, to increase geometric redundancy). Once measured, iTwin Capture Modeler computes the 3D coordinate of the point via a robust spatial intersection based on the known camera poses from the AT. As shown in Figure \ref{fig:ground_truth}, this multi-image method ensures a rigorous and reliable 3D position, which served as our ground truth for all subsequent error analysis.

The ground-truth measurements are themselves subject to operator-dependent uncertainty. However, since the same operator and protocol were used consistently across all methods, this uncertainty is expected to affect all methods similarly and does not bias the relative comparison.

\subsubsection{Comparison Method}
For a baseline comparison, we performed standard point measurements on mesh. These measurements were conducted in the iTwin Capture Desktop Viewer \cite{Bentley2024}, which provides tools to view the mesh model and measure points, lines, and surfaces by clicking directly on the 3D mesh geometry.

\subsection{Accuracy Assessment}

\begin{figure}
    \centering
    \includegraphics[width=1\linewidth]{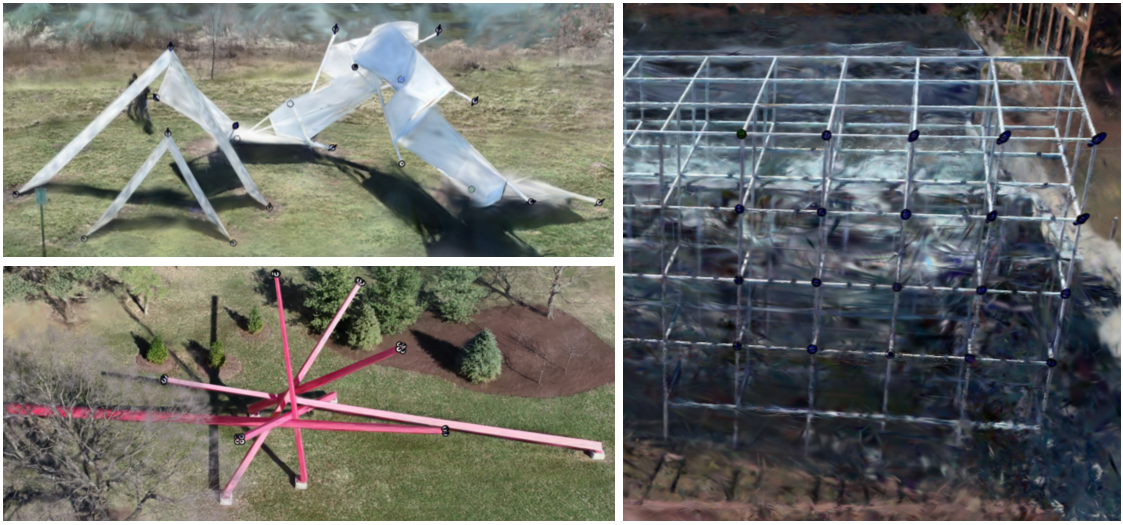}
    \caption{Distribution of the 20 validation points (VPs) used for the accuracy assessment, shown on the 3DGS rendering of the three datasets. (Top Left) Dataset 1: steel sculpture; (Bottom Left) Dataset 2: wood sculpture. (Right) Dataset3: campus building.}
    \label{fig:vps_measurement}
\end{figure}

An operator measured the 20 ground truth VPs (Figure \ref{fig:vps_measurement}) on all three datasets using two methods:
\begin{enumerate}
    \item \textbf{Mesh:} A single-click measurement on the 3D mesh model in the iTwin Capture viewer.
    \item \textbf{3DGS ($N=5$):} Our web-based method, using $N=5$ rays aimed at the target in the 3DGS rendering.
\end{enumerate}
The 3D coordinate error was computed for each VP by comparing the measured coordinate to the ground truth coordinate. The Root Mean Square Error (RMSE) was then calculated for each method.

The results, summarized in Table \ref{tab:accuracy}, clearly demonstrate that the 3DGS spatial intersection method achieved a consistently higher accuracy (lower RMSE) and better precision(lower standard deviation) than the conventional mesh-picking method across all three datasets.

This superior performance is attributable to two primary factors that directly address the limitations of mesh-picking.

First, our method improves statistical stability through redundancy. By computing a least-squares solution from $N>2$ rays, it inherently averages out and mitigates the small, random human aiming errors from any single viewpoint. The mesh-picking method, in contrast, is an $N=1$ measurement; its accuracy is entirely and precariously dependent on the perfection of a single click from a single line of sight.

Second, the 3DGS rendering provides a higher-fidelity target for aiming. Mesh models are, by nature, an \textit{approximation} of the real-world geometry, subject to surface smoothing and texture interpolation that can blur or slightly displace sharp features. The 3DGS rendering, however, provides a photorealistic visual reference that is closer to the original source imagery, allowing the operator to aim at a sharper, more defined visual target.

The "ambiguity" of the single-click method, as noted, is its fundamental flaw. A user \textit{thinks} they are clicking a feature's apex, but they are actually just intersecting a ray with a pre-existing mesh triangle. Our 3DGS method avoids this entirely: the user identifies the \textit{visual feature} from multiple angles, and the algorithm \textit{computes} its 3D coordinate. This process inherently resolves parallax ambiguity and is not constrained by a pre-computed, and potentially smoothed, mesh surface.

While the accuracy gains on these well-defined validation points are in the 1-2 cm range, this test highlights a critical weakness in mesh-based metrology. As we will demonstrate in the following section, this weakness escalates from a minor inaccuracy to a complete measurement failure when applied to difficult or thin structures.

\begin{table}[htbp]
  \centering
  \caption{Accuracy assessment (in meters) comparing mesh and 3DGS measurements against 20 validation points (VPs) per dataset. \textbf{ME} = Mean Error (Bias), \textbf{Std} = Standard Deviation (Precision).}
  \label{tab:accuracy}
  \small
  
  \begin{tabular*}{\columnwidth}{@{\extracolsep{\fill}} l l c c c c}
    \toprule
    \textbf{Dataset} & \textbf{Method} & \textbf{N} & \textbf{RMSE} & \textbf{ME} & \textbf{Std} \\
    \midrule
    \textbf{1 (Steel)} & Mesh & 20 & 0.030 & 0.028 & 0.015 \\
                       & \textbf{Ours} & 20 & \textbf{0.022} & \textbf{0.022} & \textbf{0.003} \\
    \addlinespace 
    \textbf{2 (Wood)}  & Mesh & 20 & 0.034 & 0.033 & \textbf{0.005} \\
                       & \textbf{Ours} & 20 & \textbf{0.021} & \textbf{0.019} & 0.006 \\
    \addlinespace
    \textbf{3 (Building)} & Mesh & 20 & 0.031 & 0.028 & 0.024 \\
                          & \textbf{Ours} & 20 & \textbf{0.019} & \textbf{0.014} & \textbf{0.008} \\
    \bottomrule
  \end{tabular*}
\end{table}

\subsection{Examples of measuring complex objects}

This test highlights the primary advantage of 3DGS in capturing detail where meshes fail, particularly with common UAV capture scenarios. We selected target points on two types of 'difficult' features:
\begin{enumerate}
    \item \textbf{Thin Structures (Dataset 1):} Thin, pole-like structures (e.g., traffic light poles, instruction poles) on the Statue dataset. These are classic failure cases for meshing algorithms, which tend to reconstruct them as fragmented or discontinuous.
    \item \textbf{Low-Angle/Sharp Corners (Dataset 2):} Geometrically sharp, rectangular corners at the base of the "Red Wood" statue. These features were captured from high-angle UAV-only imagery. This common capture scenario resulted in poor viewing angles and insufficient image data near the ground, causing the mesh reconstruction algorithm to fail.
\end{enumerate}
An operator attempted to measure a key point on these features using both the mesh-pick method and our 3DGS method.

The results from this test highlight the practical limitations of mesh-based metrology and the distinct advantages of our 3DGS-based method.

\begin{figure}[h]
    \centering
    \includegraphics[width=1.0\linewidth]{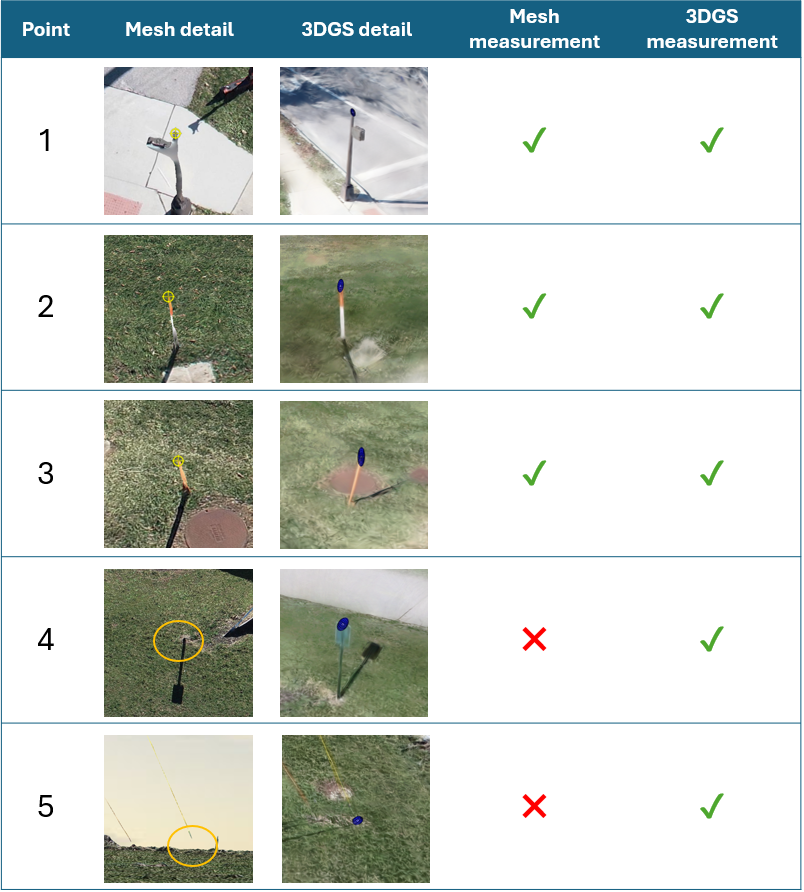}
    \caption{\textbf{Visual analysis of measurement feasibility on thin structures (Dataset 1).} This table compares measurement attempts on five different thin, pole-like features. The \textit{Mesh detail} column (Col. 2) shows the mesh geometry, which appears missing or discontinuous for points 4 and 5 (circled in yellow), leading to measurement failure. The \textit{3DGS detail} column (Col. 3) shows the visually coherent 3DGS rendering, which enabled successful and precise measurements (indicated by the blue error ellipsoid) for all five points.}
    \label{fig:Adv test 1}
\end{figure}

For \textbf{thin structures (Dataset 1)}, five feature points were selected on pole-like objects. The mesh-based method could only produce a valid measurement for three of them, as the mesh geometry for the other two points was fragmented and discontinuous, as shown in Figure \ref{fig:Adv test 1} Col. 2. The proposed 3DGS-based method, however, successfully measured all five points because the 3DGS rendering remained visually coherent (Figure \ref{fig:Adv test 1} Col. 3). Furthermore, for the three measurable points, the mesh method yielded an RMSE of \textbf{0.062 m}, while the 3DGS method achieved a significantly better RMSE of \textbf{0.037 m}. This demonstrates a clear improvement in both measurement completeness and accuracy.

\begin{figure}[h]
    \centering
    \includegraphics[width=1.0\linewidth]{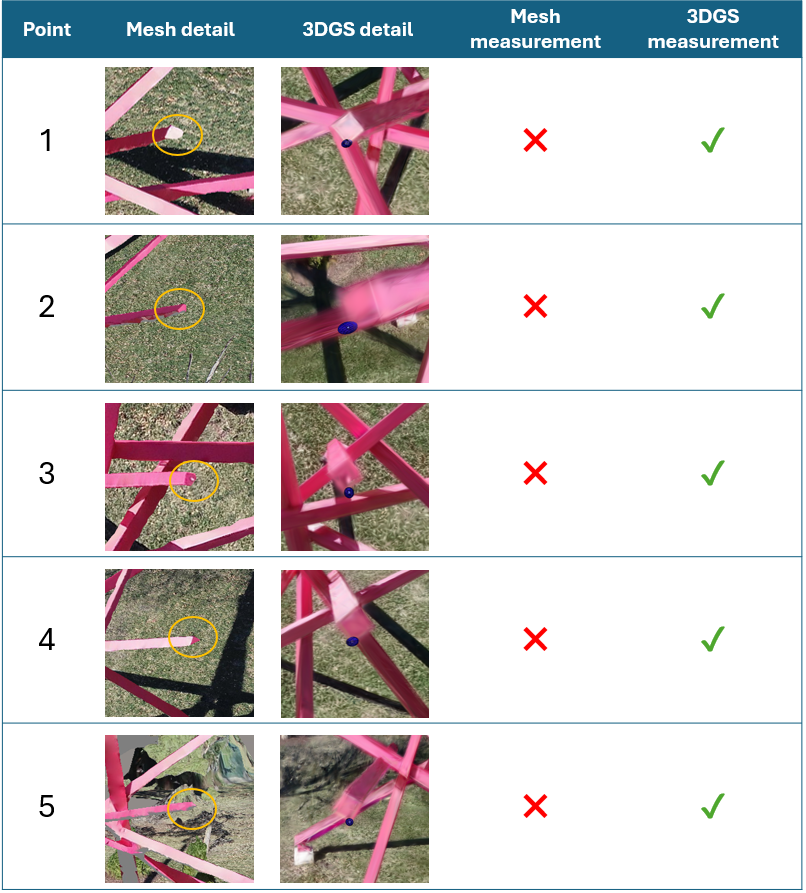}
    \caption{\textbf{Measurement feasibility on sharp corners (Dataset 2).} The mesh model (column 2) fails to reconstruct the sharp corners due to poor UAV viewing angles, making measurement impossible ("\ding{55}"). The 3DGS rendering (column 3) clearly depicts the corners, enabling successful measurement ("\ding{51}").}
    \label{fig:adv_test2}
\end{figure}

For \textbf{low-angle sharp corners (Dataset 2)}, five feature points were selected. The mesh reconstruction completely failed to recover these features due to the high-angle (nadir-only) UAV imagery, which lacked oblique views of the vertical faces (Figure \ref{fig:adv_test2} Col. 2). This resulted in \textbf{zero measurable points} for the mesh-picking method. In stark contrast, the 3DGS rendering, optimized for view synthesis, clearly depicted the corner (Figure \ref{fig:adv_test2} Col. 3). This allowed the 3DGS method to successfully measure all five points with a high-accuracy RMSE of \textbf{0.013 m}, demonstrating its robustness even with challenging capture geometries.

These two failure cases are directly linked to the fundamental assumptions of their respective technologies. Mesh reconstruction pipelines are optimized for continuous surfaces and are highly dependent on capture geometry; they often fail when faced with thin objects (which violate surface patch assumptions) or poor viewing angles (which create data gaps). Our method is immune to these geometric reconstruction failures. 3DGS is optimized for \textbf{view synthesis}, not \textbf{geometric explicitness}. As long as a feature is visible in the source images, 3DGS can typically produce a photorealistic rendering of it. Our system leverages this: it does not "pick" a model, but rather uses the high-fidelity rendering as a visual guide to perform a robust photogrammetric spatial intersection. This effectively bypasses the entire mesh-generation problem, allowing operators to measure what they can \textit{see}, not just what the meshing algorithm managed to \textit{build}. 

\section{Conclusions}
This paper introduced and validated a proof-of-concept (PoC) web application that bridges the gap between the high-fidelity rendering of 3D Gaussian Splatting (3DGS) and the critical industry need for precise 3D metrology. We demonstrated that by re-implementing the classic photogrammetric principle of spatial intersection, a 3DGS scene can serve as a robust visual reference for accurate coordinate measurement.

Our experiments confirm our primary hypothesis. First, on well-defined validation points, our multi-ray least-squares method achieves a high measurement accuracy (RMSE of 1-2 cm) that is superior to direct measurements on derivative mesh models. Second, and more significantly, our method excels where mesh-based approaches fail. The 3DGS renderings retain fine details of thin structures and sharp corners that are lost in mesh reconstruction, allowing our tool to provide accurate measurements where mesh-based methods are either highly inaccurate or entirely impossible.

This PoC highlights several key advantages of this "measurement-on-rendering" approach: (1) it democratizes metrology by removing the need for specialized stereo workstations; (2) it is accessible to all operators, not just those with biological stereo-viewing capabilities; (3) it simplifies workflows by using camera poses directly from the rendering engine; and (4) it naturally supports robust, multi-ray ($N>2$) intersections for higher precision.We acknowledge that traditional stereoscopic measurement still holds advantages, particularly in textured-poor regions where human operators can leverage structural context. Our method is presented as a powerful and accessible alternative for a wide range of applications, transforming 3DGS from a passive visualization tool into an active platform for quantitative analysis.

Future work will focus on enhancing automation. While the current method relies on manual aiming from multiple views, a clear next step is to integrate semi-automated point correspondence by leveraging feature matching techniques across renderings. Further development will also involve expanding the toolset to support other essential metrology tasks, such as measuring distances, areas, and volumes. A significant research direction will be applying this to LoD2 building model extraction, specifically by developing methods to automatically generate topological relationships (e.g., building footprints, rooflines) from a sequence of related point measurements.

\section*{Acknowledgements}

This work is supported by the Office of Naval Research (Award No. N000142312670) and Intelligence Advanced Research Projects Activity (IARPA) via Department of Interior/Interior Business Center (DOI/IBC) contract number 140D0423C0075.

{
	\begin{spacing}{1.17}
		\normalsize
		\bibliography{references,qinlab,pers_wos_2024_abs,pers_wos_2025_9} 
	\end{spacing}
}

\end{document}